# Light-YOLOv5: A Lightweight Algorithm for Improved YOLOv5 in Complex Fire Scenarios

Xu Hao [1], Li Bo [1] and Zhong Fei [1,*]

1    School of Mechanical Engineering, Hubei University of Technology, Wuhan 430000, China

**Abstract:** Fire-detection technology is of great importance for successful fire-prevention measures. Image-based fire detection is one effective method. At present, object-detection algorithms are deficient in performing detection speed and accuracy tasks when they are applied in complex fire scenarios. In this study, a lightweight fire-detection algorithm, Light-YOLOv5 (You Only Look Once version five), is presented. First, a separable vision transformer (SepViT) block is used to replace several C3 modules in the final layer of a backbone network to enhance both the contact of the backbone network to global in-formation and the extraction of flame and smoke features; second, a light bidirectional feature pyramid network (Light-BiFPN) is designed to lighten the model while improving the feature extraction and balancing speed and accuracy features during a fire-detection procedure; third, a global attention mechanism (GAM) is fused into the network to cause the model to focus more on the global dimensional features and further improve the detection accuracy of the model; and finally, the Mish activation function and SIoU loss are utilized to simultaneously increase the convergence speed and enhance the accuracy. The experimental results show that compared to the original algorithm, the mean average accuracy (mAP) of Light-YOLOv5 increases by 3.3%, the number of parameters decreases by 27.1%, and the floating point operations (FLOPs) decrease by 19.1%. The detection speed reaches 91.1 FPS, which can detect targets in complex fire scenarios in real time.

## 1 Introduction

Fires can have a major impact on public safety, and every year they cause a large number of deaths, injuries and property damage. According to incomplete statistics, China received 449,000 reported fires in the first half of 2022, resulting in 1,025 deaths, 1,001 injuries and 3.31 billion yuan in damages. Especially in complex environments such as forests and cities, once a fire spreads, it will be difficult to fight it effectively, so timely detection of the fire can greatly reduce casualties and losses.

Traditional fire detection methods mainly use smoke and temperature sensors, which have a limited detection range and scenarios, and long response times. With the development of artificial intelligence and machine learning, fire detection based on deep learning has been widely used. However, fire detection scenarios are often too complex and changeable, in this case, the generalization and robustness of traditional fire detection algorithms are not sufficient and deployment to low computing power platforms is difficult.. To address the shortcomings of existing fire detection, this paper proposes a lightweight Light-YOLOv5s algorithm for complex fire scenarios detection based on YOLOv5. The contributions of this paper are as follows:

1. Replace the last layers of the backbone network with SepViT Block, strengthen the network's connection to global feature information.
2. A Light-BiFPN structure is proposed to reduce the computational cost and parameters while enhancing the fusion of multi-scale features and enriching the semantic features.
3. We incorporate the Global Attention Mechanism into YOLOv5 to enhance the overall feature extraction capability of the network.
4. We finally verify the validity of Mish activation function and SIoU loss function.

The rest of this paper is as follows: Section 2 focuses on the work related to fire detection. Section 3 describes the framework of the model and the details of the implementation. Section 4 verifies the effectiveness of the algorithm through experiments, and Section 5 concludes with a summary.

## 2 Related Work

Due to the irregular shape of smoke, uneven spatial distribution and short existence time, it is very difficult for the accurate detection of smoke. Traditional methods usually detect smoke by features such as color, texture, and shape of smoke. Favorskaya et al.[1] used dynamic texture features to detect smoke using two-dimensional and three-dimensional LBP histograms, which can exclude the interference of wind in static scenes. Dimitropoulos et al.[2] used HSV model and adaptive median algorithm for preprocessing, and then used high-order linear dynamic system for dynamic texture analysis of smoke, which significantly improves the detection accuracy. Wang et al. [3] proposed a flame detection method that combines the dynamic and static features of the flame in the video and reduces the influence of the environment by combining flame color features and local features.

In recent years, with the rapid development of machine learning, the latest target detection algorithms in deep learning have been applied to the field of fire detection. Wang et al.[4] used a improved YOLOv4 network for real-time smoke and fire detection, dramatically reducing the number of parameters to improve detection speed and successfully deploying to UAVs, but with lower accuracy than the original algorithm. Zhang et al. [5] proposed a T-YOLOX fire detection algorithm using VIT technique to improve the accuracy of detecting smoke, fire, and people, but did not discuss the number of parameters and computational effort. Zhao et al.[6] proposed an improved Fire-YOLO algorithm for forest fires to enhance the detection of small targets in fires and reduce the model size, but again, the number of parameters and computational effort were not discussed. Li et al. [7] improved the algorithm of YOLOv3-tiny to improve the fire detection accuracy by multi-scale fusion and k-means clustering, but the detection speed is not ideal and the scenario of application is single. Yue et al.[8] reduced the false detection rate by increasing the resolution of the feature map and expanding the perceptual field, but the detection speed is not satisfactory. Wu et al. [9] added dilated convolution to the SPP module of YOLOv5 and used GELU activation function and DIoU-NMS to improve the speed and accuracy to improve the robustness of fire detection and meet the requirements of video fire detection. Zheng et al. [10] built an improved DCNN model to identify forest fires and used migration learning and PCA techniques to improve the accuracy, but did not discuss the analysis of real-time performance. Xue et al. [11] uses the SPPFP module to replace the SPPF module in YOLOv5, adds the CBAM attention module, and uses migration learning and other methods to improve the detection accuracy of small and medium-sized targets in forest fires, finally achieving good results, but sacrificing speed. Wu et al. [12] improved YOLOv4-tiny by adding SE attention mechanism and using multi-scale detection to enhance the detection of small targets and occluded objects to meet the requirements of ship fires in the sea. Shahid et al. [13] used the vision transformer for fire detection and demonstrated the feasibility of VIT, but the number of parameters and computational effort could be improved, and real-time performance was not discussed. However, some of these articles mentioned above are unsatisfactory in terms of speed, some are unsatisfactory in terms of accuracy, and some are too homogeneous in terms of environment to achieve a balance between these three. Therefore, this paper proposes a Light-YOLOv5s method for fire detection to achieve a balance of speed and accuracy in complex environments.

## 3 Methods

*3.1 Baseline*

YOLOv5 is the object detection network of the YOLO series, which is famous for being fast, lightweight and accurate. The structure of YOLOv5 consists of 4 modules are input, backbone, neck, and prediction. Compared with YOLOv4, YOLOv5 adds mosaic data enhancement and adaptive anchor frame calculation, using Leaky ReLU and Sigmoid activation functions, etc. YOLOv5 has n, s, m, l and x versions. We chose YOLOv5n, which has both speed and accuracy, as the baseline for improvement after experimental comparison, and we call the improved model Light-YOLOv5, whose structure is shown in Figure 1.

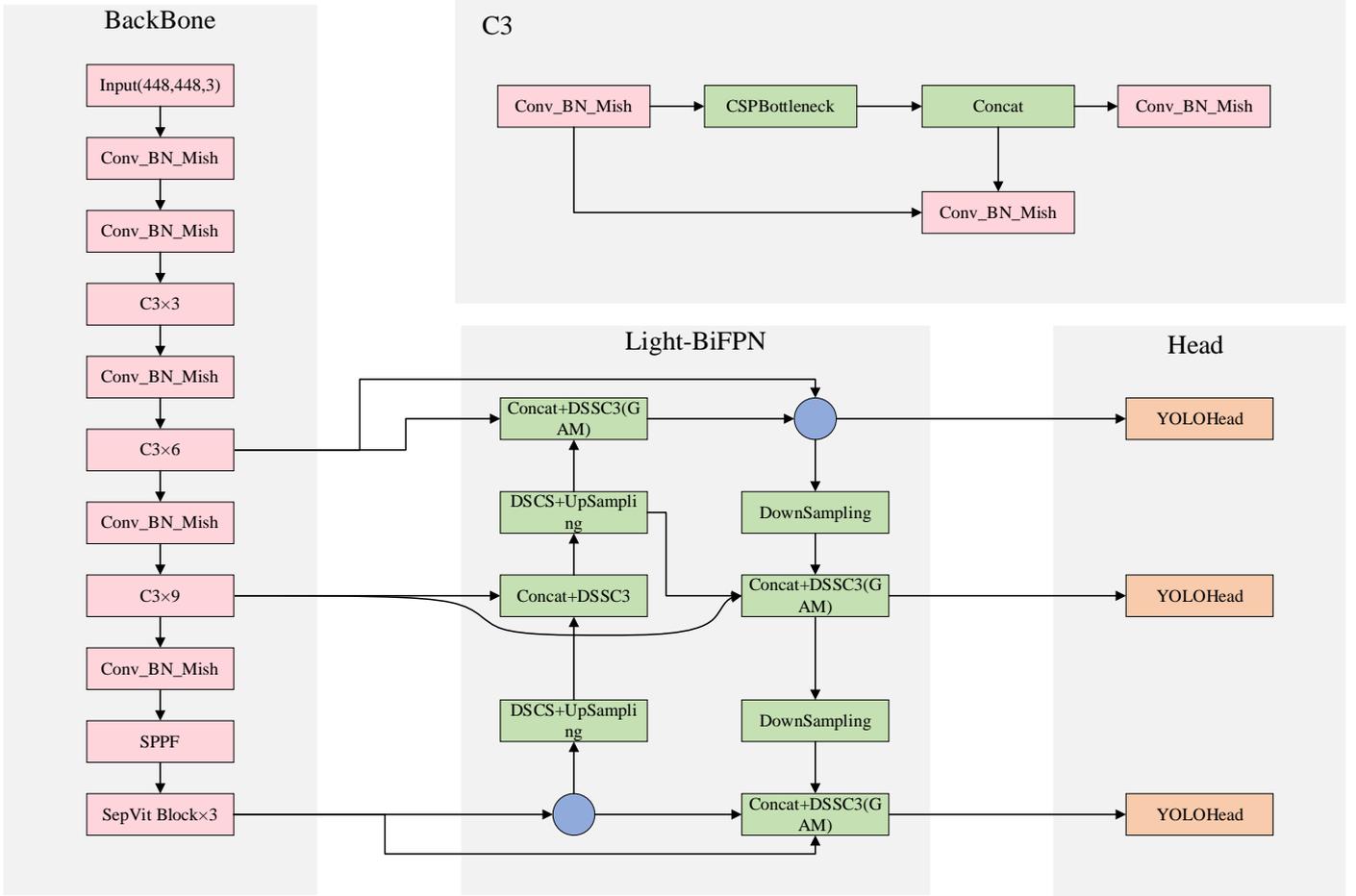

**Figure 1** Architecture of Light-YOLOv5 model

*3.2 Separable Vision Transformer*

In recent years, Vision Transformer[14-15] has achieved great success in a range of computer vision tasks, boasting performance that exceeds that of CNNs in major domains. However, these performances usually come at the cost of increased computational complexity, the number of parameters. These algorithms often require expensive GPU computing power to use and are also difficult to deploy on mobile devices.

Separable Vision Transformer[16] solves this challenge by maintaining accuracy while balancing computational cost. In this paper, the last layer of the backbone network is replaced with SepViT Block, which enhances the feature extraction capability of the model and optimizes the relationship of the global information of the network. In SepViT Block, the depthwise self-attention and pointwise self-attention to reduce computation and enable local information communication and global information interaction in windows. First, each window of the divided feature map is considered as one of its input channels, and each window contains its own information, and then a depthwise self-attention(DWA) is performed on each window token and its pixel tokens. The operation of DWA is as follows:

$$\text{DWA}(f) = \text{Attention}(f \cdot W_Q, f \cdot W_K, f \cdot W_V) \quad (1)$$

where $f$ is the feature tokens, composed of window tokens and pixel tokens. $W_Q$, $W_K$, and $W_V$ represent three Linear layers for query, key and value computation in a routine self-attention. Attention represent a standard self-attention operation. After the DWA operation is completed, pointwise self-attention(PWA) is used to establish connections among windows and generate the attention map by LayerNormalization(LN) and Gelu activation function. The operation of PWA is as follows:

$$\text{PWA}(f, wt) = \text{Attention}(\text{Gelu}(\text{LN}(wt)) \cdot W_Q, \text{Gelu}(\text{LN}(wt)) \cdot W_K, f) \quad (2)$$

where $wt$ means the window token. Then, SepViT Block can be expressed as:

$$\tilde{f}^{n} = \text{Concat}(f^{n-1}, wt) \tag{3}$$

$$\ddot{f}^{n} = \text{DWA}(\text{LN}(\tilde{f}^{n})) \tag{4}$$

$$\dot{f}^{n}, \dot{wt} = \text{Slice}(\ddot{f}^{n}) \tag{5}$$

$$\hat{f}^{n} = \text{PWA}(\dot{f}^{n}, \dot{wt}) + f^{n-1} \tag{6}$$

$$f^{n} = \text{MLP}(\text{LN}(\hat{f}^{n})) + \hat{f}^{n} \tag{7}$$

where $f^n$ represent as the SepViT Block. $\dot{f}^n$ and $\dot{wt}$ are feature maps and the learned window tokens. Concat denote the concatenation operation. Slice denote the slice operation. Figure 2 shows the structure of SepViT Block.

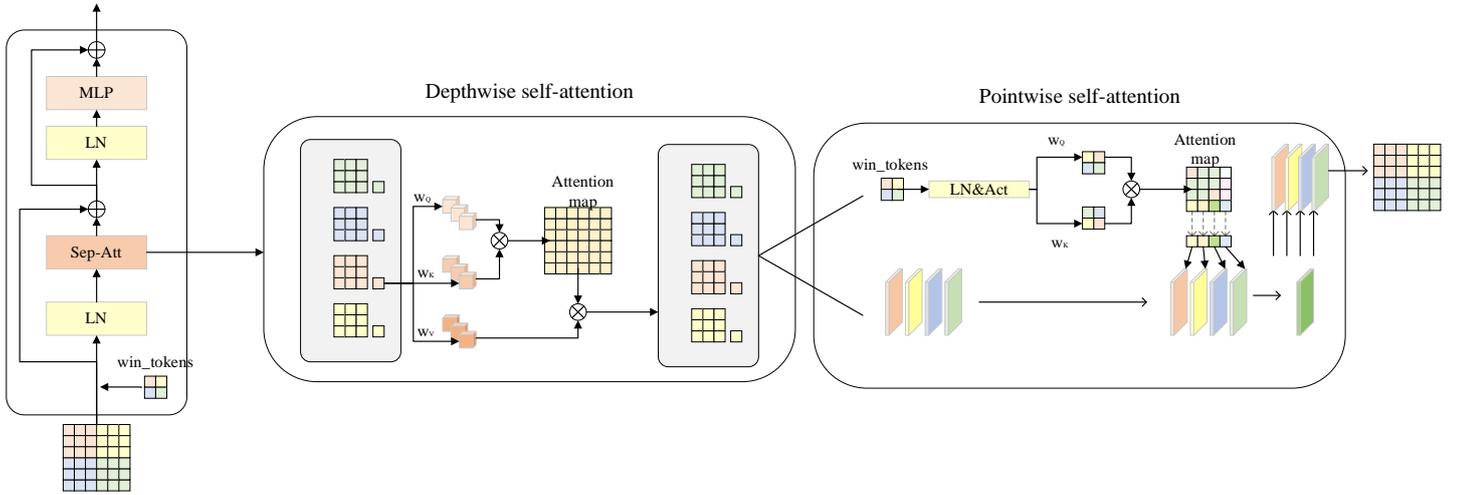

**Figure 2** Overall structure of SepViT Block

*3.3 Light-BiFPN Neck*

In this section, inspired by these articles, such as MobileNet[17-19], ShuffleNet[20-21], EfficientDet[22], GhosetNet[23], PP-LCNet[24] and so on. we have designed a lightweight neck network which we call Light- BiFPN.

In fire detection, where speed and accuracy are equally important, and timely and accurate detection of fires can greatly reduce damage. We found in our experiments that the depth-wise separable convolution(DSC) and Ghost convolution with little difference in accuracy. DSC is able to reduce the number of parameters and calculations to a greater extent, but DSC also has the drawback that the channel information of the input image is separated during the calculation. To solve this problem, we improve the DSC block in [24] by channel shuffle the features of the DSC output., and we call the improved module DSSConv, whose structure is shown in Figure 3(b). where the depth-separable convolution consists of depth convolution and point convolution. Input a $H \times W \times C$ feature map $P$, depth-wise convolution with one filter per input channel can be described as:

$$G_{k,l,m} = \sum_{i,j} K_{i,j,m} \cdot P_{k+i-1, l+j-1, m} \tag{8}$$

where $K$ is the depth-wise convolutional kernel of size $H_k \times W_k \times C$ where the $m_{th}$ filter in $K$ is applied to the $m_{th}$ channel in $P$ to produce the $m_{th}$ channel of the filtered output feature map $G$. The new features are then generated by 1×1 point convolution, The calculation process is shown in Figure 3(a).

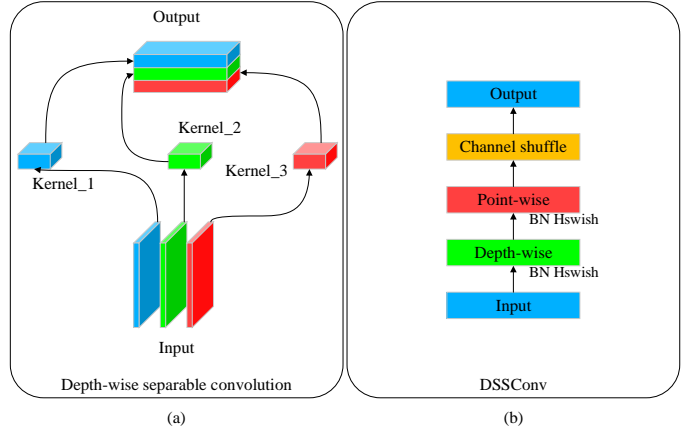

**Figure 3** (a) The calculation process of the DSC. (b)The structure of the DSSConv module.

We designed DSSbottleneck as well as DSSC3 based on the bottleneck and C3 modules of YOLOv5, and their structures are shown in (a) and (b) of Figure 4.

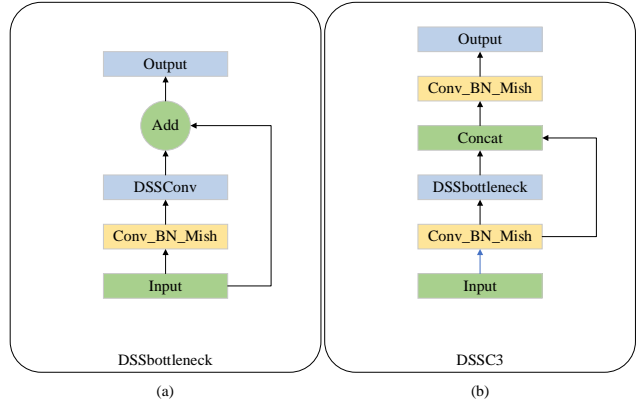

**Figure 4** (a)The structure of the DSSbottleneck module. (b)The structures of the DSSC3 module.

YOLOv5 uses PANET[25] on the neck for feature extraction and fusion. It uses bottom-up and top-down bidirectional fusion methods and achieves good results, but the environment of fire detection is usually too complex and more features need to be fused to get better results. BiFPN is a weighted bidirectional feature pyramid network that connects input and output nodes of the same layer across layers to achieve higher-level fusion and shorten the path of information transfer between higher and lower layers. Since weighting brings a certain computational rise, this paper removes the weighted feature fusion to make the neck network further lightweight.

*3.4 Global Attention Mechanism*

The complex environment of fire detection is prone to false and missed detections. GAM[26] is used to strengthen the connection between space and channels, reduce the information reduction of flame and smoke in fire and amplify the features of global dimension. Given an input feature map $F_1 \in R^{H \times W \times C}$, the output $F_3$ is defined as:

$$F_3 = M_S(M_c(F_1) \otimes F_1) \otimes (M_c(F_1) \otimes F_1) \tag{9}$$

Where $M_c$ is the channel map and $M_S$ is the spatial maps; $\otimes$ means element wise multiplication. We add GAM to the bottleneck module, whose structure is shown in Figure 5.

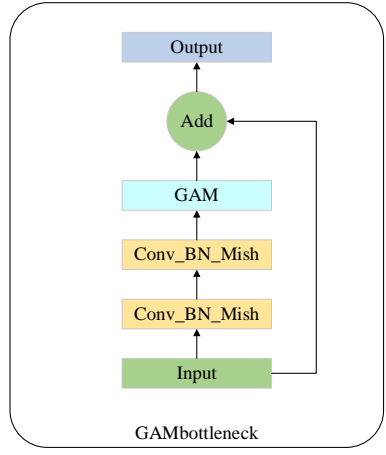

**Figure 5** The structure of the GAMbottleneck.

*3.5 The IoU Loss and Activation*

IoU[27] loss can predict the localization of the bounding box regression more accurately, and the most commonly used in the YOLO series is CIoU[28]. as the research progresses, there are more and more variants of IoU, such as DIoU[29], GIoU[30], EIoU[31] and the latest SIoU[32]. they are defined as follows:

$$Loss_{IoU} = 1 - IoU, IoU = \left|\frac{A \cap B}{A \cup B}\right| \tag{10}$$

$$Loss_{GIoU} = 1 - IoU + \frac{C - (A \cup B)}{C} \tag{11}$$

$$Loss_{DIoU} = 1 - IoU + \frac{\rho^2_{(b,b^{gt})}}{d^2} \tag{12}$$

$$Loss_{CIoU} = 1 - IoU + \frac{\rho^2_{(b,b^{gt})}}{d^2} + \alpha v,$$

$$\alpha = \frac{v}{(1 - IoU) + v}, \tag{13}$$

$$v = \frac{4}{\pi^2}(\arctan\frac{w^{gt}}{h^{gt}} - \arctan\frac{w}{h})^2$$

$$Loss_{EIoU} = 1 - IoU + \frac{\rho^2_{(b,b^{gt})}}{d^2} + \frac{\rho^2_{(w,w^{gt})}}{C_w^2} + \frac{\rho^2_{(h,h^{gt})}}{C_h^2} \tag{14}$$

$$Loss_{SIou} = 1 - IoU + \frac{\Delta + \Omega}{2},$$

$$\Delta = \sum_{t=x,y}(1-e^{-\gamma\rho_t}),$$

$$\rho_x = \frac{b_{c_x}^{gt} - b_{c_x}}{c_w}, \rho_y = \frac{b_{c_y}^{gt} - b_{c_y}}{c_h}, \gamma = 2 - \Lambda,$$

$$\Lambda = 1 - 2*\sin^2(\arcsin(x) - \frac{\pi}{4}),$$

$$x = \frac{c_h}{\sigma} = \sin(\alpha),$$

$$\sigma = \sqrt{\left(b_{c_x}^{gt} - b_{c_x}\right)^2 + \left(b_{c_y}^{gt} - b_{c_y}\right)^2},$$

$$c_h = \max(b_{c_y}^{gt}, b_{c_y}) - \min(b_{c_y}^{gt}, b_{c_y}),$$

$$\Omega = \sum_{t=w,h}(1-e^{-\omega_t})^\theta,$$

$$\omega_w = \frac{|w - w^{gt}|}{\max(w, w^{gt})}, \omega_h = \frac{|h - h^{gt}|}{\max(h, h^{gt})} \tag{15}$$

where the parameters $A$ and $B$ represent the area of the ground truth bounding box and the area of the prediction bounding box, respectively; $C$ denotes the minimum enclosing box of the ground truth bounding box and the prediction bounding box; $b, b^{gt}$ represent the centroids of the prediction bounding box and the ground truth bounding box, respectively, and $\rho$ represents the Euclidean distance between the two centroids, $d$ is the diagonal distance of the smallest enclosing region that can contain both the prediction bounding box and the ground truth bounding box; $\alpha$ is the weight function, and $v$ is used to measure the similarity of aspect ratio.

The CIoU used by YOLOv5 relies on the aggregation of bounding box regression metrics and does not consider the direction of the mismatch between the desired ground box and the predicted "experimental" box. This leads to inferiority to SIoU in terms of training speed and prediction accuracy.

In lightweight networks, HSwish, Mish and LeakyReLu are faster than ReLu in terms of training speed. They can be defined as:

$$H-swish(x) = x\frac{\text{ReLu6}(x+3)}{6} \tag{16}$$

$$Mish(x) = x \cdot \tanh(\log(1+e^x)) \tag{17}$$

$$LeakyReLu(x) = max(ax, x) \tag{18}$$

We found experimentally that using the Mish activation function is more accurate than the others, and detailed comparison experiments are given in Section 4.

**4 Experiment**

*4.1 Datasets*

Since there is a lack of authoritative datasets for fire detection, the dataset used in this paper is derived from public datasets and web images and contains 21136 images. The dataset we collected contains various scenarios, such as forest fires, indoor fires, urban fires, traffic fires, etc. Figure 6 shows part of the dataset.

*4.2 Training Environment and Details*

This paper uses Ubuntu 18.04 operating system, NVIDIA GeForce RTX3060 GPU, CUDA11.1, Python3.8.8, PyTorch1.8.0. We randomly divide the dataset into training data, validation data and test data by 8:1:1 and use SGD optimizer for training, with 16 Batchsize, initial learning rate of 0.01, and 100 training epochs, the size of the input image is 448×448.

*4.3 Model Evaluation*

In this paper, precision(P), recall(R), average precision(AP), mean average precision(mAP), parameters, computation, inference time, and FPS are used as evaluation metrics for model performance. where AP is the area under the PR curve and mAP denotes the average of AP for each category. The specific formula is as follows:

$$P=\frac{TP}{TP+FP} \quad (19)$$

$$R=\frac{TP}{TP+FN} \quad (20)$$

TP (True Positives) means that the sample is divided into positive samples, and is divided correctly. FP (False Positives) means that the sample is divided into positive samples, and it is divided into the incorrectly samples. FN (False Negatives) means that the sample is divided into negative samples, and it is divided into the incorrectly samples.

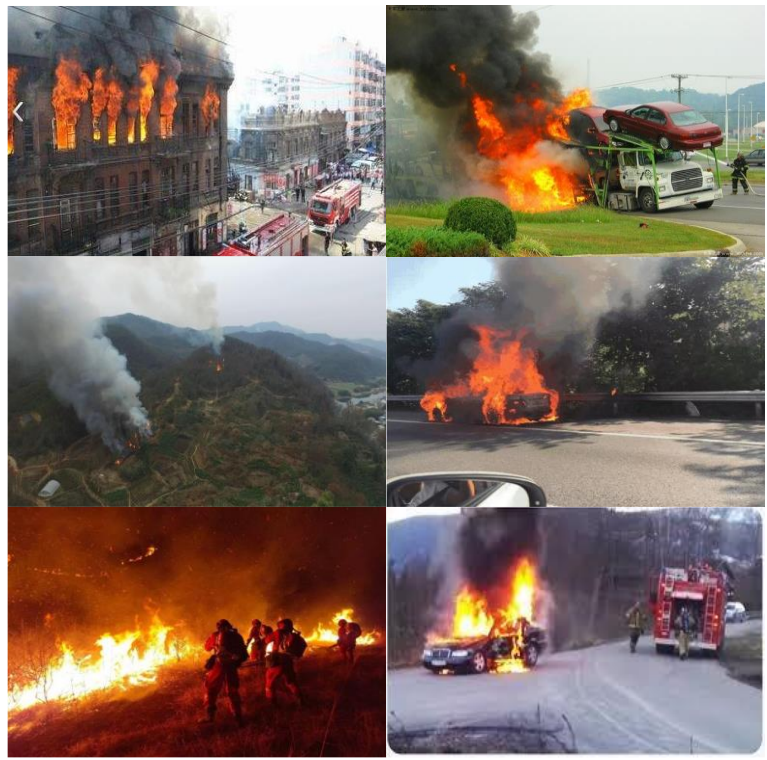

**Figure 6** Example images of the dataset

*4.4 Result Analysis and Ablation Experiments*

To further verify the validity of the model, in this section we perform a series of ablation experiments. As shown in Table 1, we compared the effects of different versions of YOLOv5. The mAP of YOLOv5n is 1.7% lower than YOLOv5s, 2.8% lower than YOLOv5m, 73.6% less computation than YOLOv5s, 91.3% less than YOLOv5m, 74.8% less number of parameters than YOLOv5s, 91.5% less than YOLOv5m, and faster detection speed than YOLOv5s and YOLOv5m. So we choose YOLOv5n as the baseline.

Further, we conducted comparison experiments on LeakyReLu, Mish, HSwish activation functions and CIoU, SIoU with YOLOv5n as the baseline. As shown in Table 2, The detection speed of Mish is not as fast as LeakyReLu and HSwish, but the accuracy is better, while SIoU and CIoU are about the same in detection speed and better in accuracy than CIoU, so we choose the combination of Mish activation function and SIoU loss.

To further validate the effectiveness of Light-BiFPN, we use the latest method of replacing the backbone with a lightweight network to conduct comparative experiments. We found that ShuffleNetv2 has the fastest detection speed but is hardly satisfying in terms of accuracy, while Light-BiFPN has the second fastest detection speed after ShuffleNetv2 and much higher accuracy than other lightweight networks. The results are shown in Table 3.

Finally, we compared all the improved methods for ablation experiments, and the results are shown in Table 4.

Compared with the original algorithm, the mAP of Light-YOLOv5 has been improved by 3.3%, and the number of parameters and computation have been reduced to a certain extent, although the detection speed is lower than the original algorithm, but it can also meet the industrial detection needs. We also compared with the most advanced detectors at this stage to further verify the effectiveness of the methods, and the comparison results are shown in Table 5. It can be seen that although Light-YOLOv5 is inferior to YOLOv7-tiny and YOLOv3-tiny in terms of detection speed, it is much higher than these detectors in other parameters, with mAP 6.8% higher than the latest YOLOV7-tiny, further proving the effectiveness of the method in this paper. We have placed the detection effect graph in Figure 7(b).

Table 1 Performance comparison of different models of YOLOv5

| Model | Params(M) | FLOPS(B) | mAP@0.5(%) | FPS |
|---|---|---|---|---|
| YOLOv5n | 1.77 | 4.2 | 67.6 | 111.1 |
| YOLOv5s | 7.02 | 15.9 | 69.3 | 100.0 |
| YOLOv5m | 20.87 | 48.0 | 70.4 | 87.78 |

Table 2 The comparison results of different activation functions and IoU loss under the same model

| Model | Activation/IoU Loss | Params(M) | mAP@0.5(%) | FPS |
|---|---|---|---|---|
| YOLOv5n | LeakyReLu/CIoU | 1.77 | 67.8 | 107.3 |
| | HSwish/CIoU | 1.77 | 67.3 | 109.2 |
| | Mish/CIoU | 1.77 | 68.0 | 95.3 |
| | LeakyReLu/SIoU | 1.77 | 68.3 | 107.5 |
| | HSwish/SIoU | 1.77 | 67.6 | 109.4 |
| | Mish/SIoU | 1.77 | 68.7 | 95.6 |

Table 3 The comparative experiments of Light-BiFPN and different state-of-the-art lightweight models

| Model | Params(M) | FLOPS(B) | mAP@0.5(%) | FPS |
|---|---|---|---|---|
| MobileNetv3-YOLOv5n | 1.93 | 3.5 | 62.8 | 98.2 |
| ShuffleNetv2-YOLOv5n | 0.71 | 1.0 | 61.8 | 126.3 |
| GhostNet-YOLOv5n | 1.39 | 3.3 | 64.8 | 116.4 |
| PPLCNet-YOLOv5n | 0.95 | 2.0 | 63.5 | 112.5 |
| Light-BiFPN-YOLOv5n | 1.25 | 3.3 | 68.6 | 125.6 |

Table 4 Results of ablation experiments with different modified methods

| Model | Params(M) | FLOPS(B) | mAP@0.5(%) | FPS |
|---|---|---|---|---|
| Baseline(YOLOv5n) | 1.77 | 4.2 | 67.6 | 111.1 |
| Baseline+Light-BiFPN | 1.25 | 3.3 | 68.6 | 128.6 |
| Baseline+Light-BiFPN+SepViT | 1.26 | 3.3 | 69.8 | 120.4 |
| Baseline+Light-BiFPN+SepViT+GAM | 1.29 | 3.4 | 70.3 | 106.5 |
| Baseline+Light-BiFPN+SepViT,Mish,SIou | 1.29 | 3.4 | 70.9 | 91.1 |

Table 5 Comparison of the results of the most advanced detectors at this stage

| Model | Params(M) | FLOPS(B) | mAP@0.5(%) | FPS |
|---|---|---|---|---|
| YOLOv3-tiny | 8.67 | 12.9 | 64.8 | 201.5 |
| YOLOX-s | 8.93 | 26.8 | 65.4 | 64.6 |
| YOLOv7-tiny | 6.01 | 13.1 | 64.1 | 285.3 |
| Light-YOLOv5 | 1.29 | 3.4 | 70.9 | 91.1 |

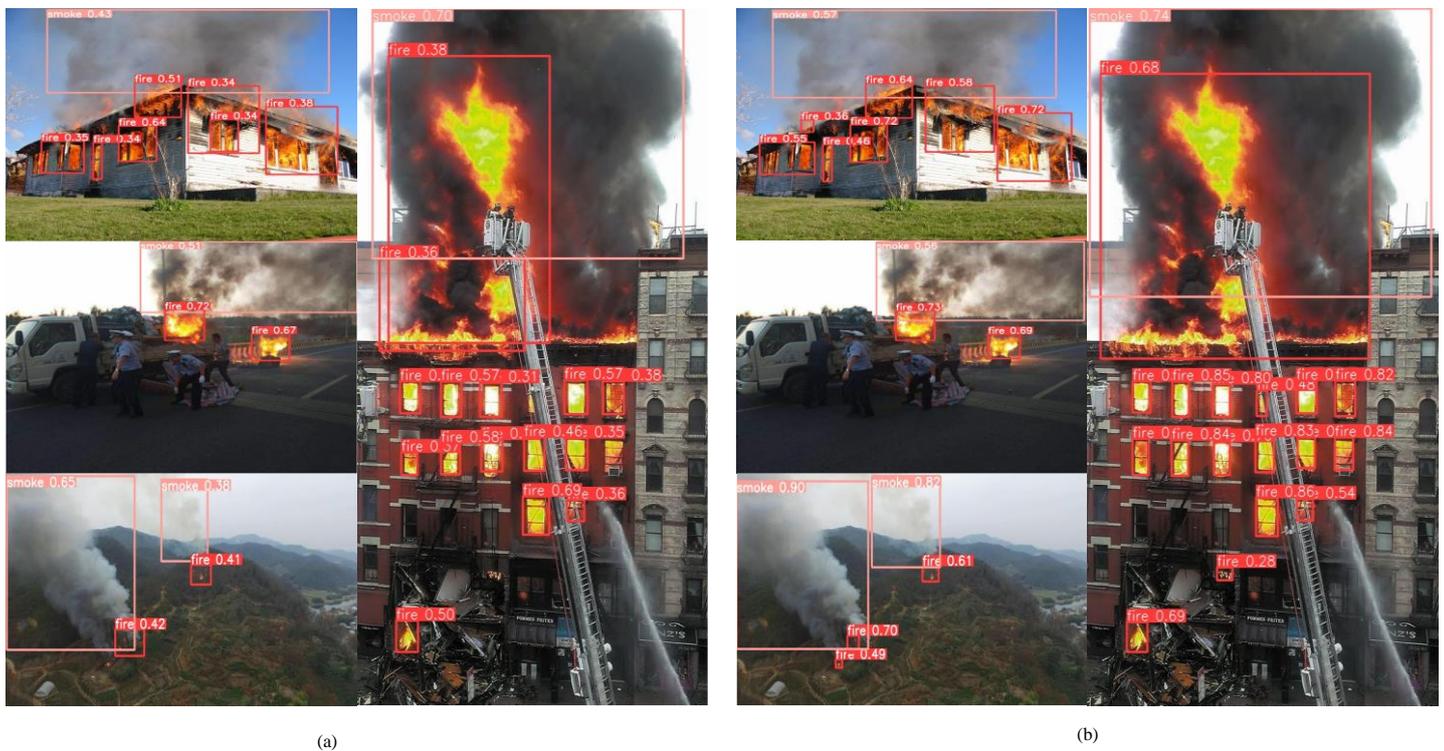

(a)　　　　　　　　　　　　　　　　　　　(b)

**Figure 7** Comparison of Light-YOLOv5 and the original algorithm detection results. (a)YOLOv5n (b)Light-YOLOv5

## 5 Discussion

In summary, this paper proposes a Light-YOLOv5 algorithm for fire detection in complex scenarios, achieving a balance of efficiency and performance. Light-YOLOv5 uses YOLOv5n as the baseline, SepViT Block to strengthen the connection between the backbone network and the global information, Light-BiFPN to strengthen the network feature extraction and lighten the network, fusion GAM module to reduce the loss of information, and finally Mish activation function to improve the accuracy and SIoU to improve the convergence speed of training. Experimental results on a self-developed fire dataset show that Light-YOLOv5 has a 3.3% higher mAP than the baseline model and 6.8% higher mAP than the latest YOLOv7-tiny compared to the state-of-the-art detector, with detection speeds that meet industrial requirements.

**Funding:** This research was sponsored by the National Natural Science Foundation of China(Grant No.52005168), Initial Scientific Research Foundation of Hubei University of Technology (BSQD2020005), and the Open Foundation of Hubei Key Lab of Modern Manufacture Quality Engineering (KFJJ-2020009).

**Conflicts of Interest:** The authors declare no conflict of interest.